\documentclass[conference]{IEEEtran}

\usepackage{times}
\usepackage{soul}
\usepackage{helvet}
\usepackage{courier}
\usepackage{array}
\usepackage{booktabs} 
\usepackage{multirow}
\usepackage[english]{babel}
\usepackage[hidelinks]{hyperref}
\usepackage[linesnumbered,ruled,vlined]{algorithm2e}
\usepackage{algpseudocode}
\usepackage{subfigure}
\usepackage{graphicx, caption}
\usepackage{amsmath}
\usepackage{bm}
\usepackage{url}
\usepackage{stmaryrd}
\usepackage{balance}
\usepackage{amssymb}
\usepackage{pgfplots}
\usepackage{pifont}
\usepackage{epsfig}
\usepackage{booktabs}
\usepackage{pgffor}
\usepackage{tikz}
\usepackage{multirow}
\usepackage{amsmath}
\usepackage{tikz}
\usetikzlibrary{positioning,arrows,calc}
\usepackage[all]{xy}
\usepackage{makecell}
\usepackage{enumitem}
\usepackage[flushleft]{threeparttable}

\usepackage{color}

\newcommand{\our}{\textsl{HDGI}}

\newtheorem{defn}{Definition}[section]\usepackage[ruled]{algorithm2e}

\newcommand{\concatenate}{\operatornamewithlimits{\|}}

\begin{document}
	
	\title{Heterogeneous Deep Graph Infomax}	
	
	\author{Yuxiang Ren,\textsuperscript{1}
		Bo Liu,\textsuperscript{2}
		Chao Huang,\textsuperscript{2}
		Peng Dai,\textsuperscript{2}
		Liefeng Bo,\textsuperscript{2}
		Jiawei Zhang,\textsuperscript{1}\\
		\textsuperscript{1}{Florida State University, IFM Lab}\\
		\textsuperscript{2}{JD Finance America Corporation, AI lab}\\
		yuxiang@ifmlab.org,
		kfliubo@gmail.com,
		chuang7@nd.edu,
		peng.dai@jd.com,
		liefeng.bo@jd.com,
		jiawei@ifmlab.org}
	\maketitle	
	\begin{abstract}
		Graph representation learning is to learn universal node representations that preserve both node attributes and structural information. The derived node representations can serve various downstream tasks, such as node classification and node clustering. 
		When a graph is heterogeneous, the problem becomes more challenging than the homogeneous graph representation learning problem. Inspired by emerging mutual information-based learning algorithms, in this paper we propose an unsupervised graph neural network \textbf{H}eterogeneous \textbf{D}eep \textbf{G}raph \textbf{I}nfomax ({\our}) for heterogeneous graph representation learning. We use the meta-path to model the structure involving semantics in heterogeneous graphs and utilize graph convolution module and semantic-level attention mechanism to capture individual node local representations. By maximizing the local-global mutual information, {\our} effectively learns node representations based on the diverse information in heterogeneous graphs. 
		Experiments show that {\our} remarkably outperforms state-of-the-art unsupervised graph representation learning methods on both node classification and clustering tasks. By feeding the learned representations into a parametric model, such as logistic regression, 
		we even achieve comparable performance in node classification tasks when comparing with state-of-the-art supervised end-to-end GNN models.

	\end{abstract}
	
	\maketitle

	\section{Introduction}
	\label{sec:intro}
	Numerous real-world applications, such as social networks~\cite{JZhang19}, financial platforms~\cite{ren2019ensemfdet} and knowledge graphs~\cite{WMWG17} exhibit the favorable property of graph structure data. Meanwhile, handling graph data is very challenging. Because each node has its unique attributes, and the connections between nodes convey essential information. When learning from nodes' attributes and the connection information simultaneously, the task becomes more challenging. 
	
	Traditional machine learning methods focus on individual nodes' features but ignore the structural information, which obstructs their ability to process graph data. Graph neural networks (GNNs) learn nodes' new feature vectors through a recursive neighborhood aggregation scheme~\cite{XHLJ10}. 
	With the support of sufficient training samples, a rich body of supervised graph neural network models have been developed~\cite{KW17,VCCRLB18,YYRHL18}. However, labeled data is not always available in graph representation learning tasks. Those algorithms are not applicable to the unsupervised learning settings. For the purpose of alleviating the training sample scarcity problem, unsupervised graph representation learning has aroused extensive research interest. The goal of this task is to learn a low-dimensional representation of each graph node. The representation preserves graph topological structure and node content. Meanwhile, the learned representations can be applied to conventional sample-based machine learning algorithms.  
	\begin{figure}[t]
		\centering
		\begin{minipage}[l]{0.8\columnwidth}
			\centering
			\includegraphics[width=\textwidth]{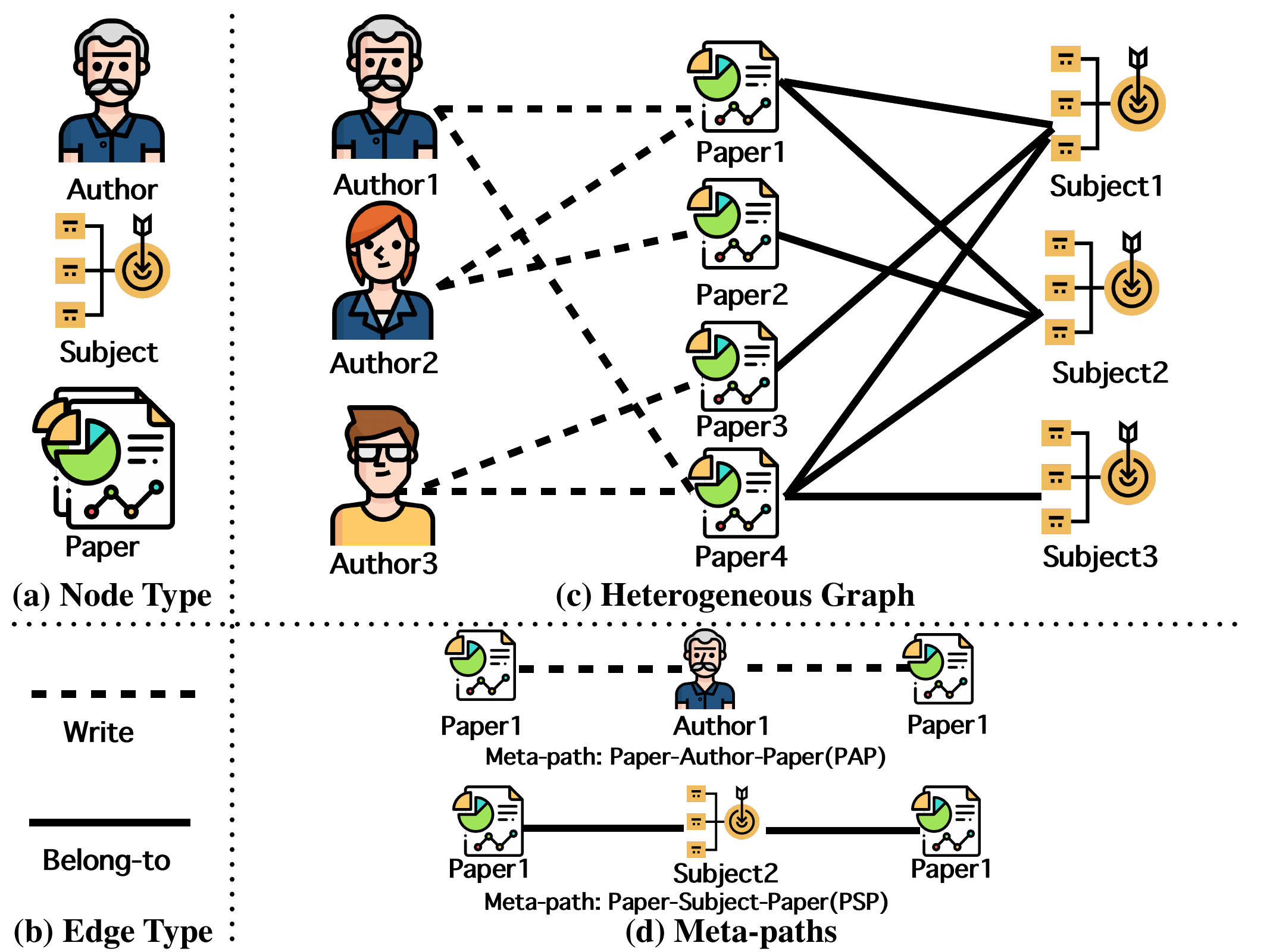}
		\end{minipage}
		\caption{An example of heterogeneous bibliographic graph.}\label{fig:acmhg}
		\vspace{-20pt}
	\end{figure}
	
	Most of the existing unsupervised graph representation learning models can be grouped into factorization-based models and edge-based models. Precisely, factorization-based models capture the global graph information by factorizing the sample affinity matrix~\cite{zhang2016collective,yang2015network,zhang2016collective}. Through the affinity matrix factorization, the full graph structure can be well grasped. However, those methods ignore the node attributes and local neighborhood relationships. Edge-based models exploit the local and higher-order neighborhood information by edge connections or random-walk paths. Nodes tend to have similar representations if they are connected or co-occur in the same path~\cite{KW16,DN17,HYL17,PAS14}. Edge-based models are prone to preserve limited order node proximity. They lack a mechanism to preserve the global graph structure. The recently proposed mutual information-based models~\cite{HFMGBTB19,velivckovic2018deep,sun2019infograph} demonstrate that mutual information maximization has attractive advantages in encouraging an encoder to considers both global and local features. For graph structure data, global and local graph structure also need to be comprehensively considered. For homogeneous graphs, DGI~\cite{velivckovic2018deep} maximizes the mutual information between graph patch representations and the corresponding high-level summaries of graphs. 
	However, the networked data in the real world usually contain complex structures (involving multiple types of nodes and edges), which can be formally modeled as heterogeneous graphs (HG). Compared with homogeneous graphs, heterogeneous graphs contain more detailed information and rich semantics among multi-typed nodes. Taking the bibliographic network in Figure~\ref{fig:acmhg} as an example, it contains three types of nodes (Author, Paper, and Subject) as well as two types of edges (Write and Belong-to). Besides, the individual nodes themselves also carry much attribute information (e.g., paper textual contents). Due to the diversity of node and edge types, the heterogeneous graph itself becomes more complicated. The diverse (direct or indirect) connections between nodes also convey more semantic information. The models initially proposed for homogeneous graphs may encounter significant challenges to handle relations with different semantics in heterogeneous graphs. 
	
	In this paper, we explore the use of mutual information maximization for heterogeneous graph representation learning. To address the aforementioned challenges in heterogeneous graphs, we propose a novel meta-path based unsupervised graph neural network model, namely \textbf{H}eterogeneous \textbf{D}eep \textbf{G}raph \textbf{I}nfomax ({\our}). In heterogeneous graph studies, meta-path~\cite{SHYYW11} has been widely used to represent the composite relations with different semantics. As illustrated in Figure~\ref{fig:acmhg}(d), the relations between paper can be expressed by PAP and PSP, which represent papers written by the same author and papers belonging to the same subject, respectively. 
	{\our} utilizes the structure of meta-paths to model connection semantics in heterogeneous graphs. Based on different meta-paths, HDGI disassembles heterogeneous graphs into homogeneous graphs of specific semantics. In these homogeneous graphs, HDGI applies a convolutional style GNN to capture the local representation of nodes with specific semantics. After that, through a semantic-level attention mechanism, HDGI aggregates node representations of different semantics. By maximizing local-global mutual information, {\our} learns high-level representations containing graph-level structural information without any supervised label. The node attributes can be simultaneously fused into the representations in the process of maximizing local-global mutual information. In order to verify the effectiveness and competitiveness of learned representations, we perform experiments on both classification tasks and clustering tasks. The experimental results show that the performance of {\our} can often beat supervised state-of-art comparative graph neural network models.

	In summary, our contributions in this paper can be summarized as follows:
	\begin{itemize}[leftmargin=*]
		\item This paper presents the first model to apply mutual information maximization to representation learning in heterogeneous graphs.
		\item Our proposed method, {\our}, is a novel unsupervised graph neural network. It handles graph heterogeneity by utilizing an attention mechanism on meta-paths and deals with the unsupervised settings by applying mutual information maximization.
		\item Our experiments demonstrate that the representations learned by {\our} are effective for both node classification and clustering tasks. Moreover, its performance can also beat state-of-the-art GNN models, where they have additional supervised label information.
	\end{itemize}
	The rest of this paper is organized as follows. We discuss the related work in Section 2. In Section 3, we present the problem formulation along with important terminologies used in our method. We present {\our} in detail in Section 4. Experimental results and analyses are provided in Section 5. Finally, we conclude the paper in Section 6.
	\section{Related Work}
	\label{sec:relate}

	\subsection{Graph representation learning} 
	Graph representation learning has become a non-trivial topic~\cite{CWPZ18} because of the ubiquity of graphs in real-world scenarios. As a data type containing rich structural information, many models~\cite{PAS14,GL16,TQWZYM15,MPJZW16,WCWPZY17} acting on graphs learn the representations of nodes based on the structure of the graph. Node2vec~\cite{GL16} learns a mapping of nodes to a low-dimensional space of features that maximizes the likelihood of preserving network neighborhoods of nodes.
	DeepWalk~\cite{PAS14} uses the set of random walks over the graph in SkipGram to learn node embeddings. Tang et al.~\cite{TQWZYM15} optimize a carefully designed objective function that preserves both the local and global network structures.
	Qu et al. ~\cite{MPJZW16} propose preserving asymmetric transitivity by approximating high-order proximity based on asymmetric transitivity. Wang et al.~\cite{WCWPZY17} attempt to retrieve structural information through matrix factorization incorporating the community structure. However, all the above methods are proposed for homogeneous graphs.
	
	In order to handle graph heterogeneity, metapath2vec~\cite{DCS17} samples random walks under the guidance of meta-paths and learns node embeddings through the skip-gram. HIN2Vec~\cite{FCL17} learns the embedding vectors of nodes and meta-paths simultaneously while conducts prediction tasks. Wang et al. \cite{WJSWCYY19} consider the attention mechanism in heterogeneous graph learning through the model HAN, where information from multiple meta-path defined connections can be learned effectively. 
	From the perspective of attributed graphs, SHNE~\cite{ZSC19} captures both structural closeness and unstructured semantic relations through joint optimization of heterogeneous SkipGram and deep semantic encoding.
	HGAT~\cite{ren2020hgat} employs a hierarchical attention mechanism considering both node-level and schema-level attention to handle graph heterogeneity. Many learning methods~\cite{schlichtkrull2018modeling,WMWG17} on knowledge graphs can often be applied to other heterogeneous graphs.
	
	\subsection{Graph neural network} 
	The core idea of GNN is aggregating the neighbors' feature information through neural networks to learn the new features~\cite{XHLJ10}, which combine the independent information of the node and corresponding structural information in the graph. A propagation model incorporating gated recurrent units to propagate information across all nodes is proposed in \cite{YDMR16}. Joan Bruna et al. \cite{JWAY13} extends convolution to general graphs by a novel Fourier transformation in graphs. Kipf et al. \cite{KW17} propose Graph Convolutional Network (GCN). Hamilton et al. \cite{HYL17} introduce GraphSAGE that generates embeddings by directly aggregating features from a node's local neighborhood. Graph Attention Network (GAT) \cite{VCCRLB18} first imports the attention mechanism into graphs. Other successful GNNs are also based on supervised learning including SplineCNN~\cite{fey2018splinecnn}, AdaGCN~\cite{sun2019adagcn} and AS-GCN~\cite{huang2018adaptive}. The unsupervised learning GNNs can be mainly divided into two categories, i.e., random walk-based~\cite{PAS14,GL16,KW16,DN17,HYL17} and mutual information-based~\cite{velivckovic2018deep}. DGI~\cite{velivckovic2018deep} is a general GNN for learning node representations within graph-structured data in an unsupervised manner. DGI relies on maximizing mutual information between patch representations and corresponding high-level summaries of graphs. DGI is the pioneer of our work, but it can only be applied to homogeneous graphs. {\our} extends mutual information-based GNNs to heterogeneous graphs.
	\subsection{Mutual information} 
	Information theory~\cite{shannon1948mathematical} has been applied in various domains. Recently mutual information is utilized in various learning tasks successfully. Belghazi et al.~\cite{BBROBCD18} propose a neural network-based mutual information estimator, enabling the mutual information estimation to be linearly scalable in dimensionality. Deep InfoMax (DIM)~\cite{HFMGBTB19} investigates unsupervised learning of representations by maximizing mutual information in the computer vision domain. DGI~\cite{velivckovic2018deep} extends the mechanism into the graph domain and works on node representation learning. Sun et al.~\cite{sun2019infograph} take advantage of mutual information maximization to learn graph-level representations.
	\section{Problem Formulation}
	\label{sec:model}
	In this section, we first introduce the concept of \textit{heterogeneous graph} and the problem definition of \textit{heterogeneous graph representation learning}. Next, we define \textit{meta-path based adjacency matrix}, which is critical in the following algorithm description.
	
	\noindent
	\begin{defn}[Heterogeneous Graph (HG)]
		A heterogeneous graph is defined as  $\mathcal{G} = (\mathcal{V}, \mathcal{E})$ with a node type mapping function $\phi : \mathcal{V}  \rightarrow \mathcal{T}$ and an edge type
		mapping function $\psi : \mathcal{E}  \rightarrow \mathcal{R}$. Each node $v \in\mathcal{V}$ belongs to one particular type in the node type set $\mathcal{T}: \phi(v) \in \mathcal{T}$, and each edge $e \in\mathcal{E}$  belongs to a particular edge type in the edge type set $\mathcal{R}: \psi(e) \in \mathcal{R}$. Heterogeneous graphs have the property that $|\mathcal{T}|+|\mathcal{R}| > 2$. The attributes and content of nodes can be encoded as initial feature matrix $X$.
	\end{defn}
	
	\noindent
	\emph{Problem Definition}. (\textit{Heterogeneous Graph Representation Learning}):  
	Given a heterogeneous graph $\mathcal{G}$ and the initial feature matrix $X$, the representation learning task in $\mathcal{G}$ is to learn the low dimensional node representations $H\in \mathbb{R}^{|\mathcal{V}|\times d}$ which contains both structure information of $\mathcal{G}$ and node attributes of $X$. The learned representation $H$ can be applied to downstream graph-related tasks such as node classification and node clustering, etc. 
	We focus on learning the representation of a specific type of node in this paper.
	We represent the set of nodes for representation learning as the target-type nodes $\mathcal{V}_t$.
	
	In a heterogeneous graph, two neighboring nodes can be connected by different types of edges. Meta-paths~\cite{SHYYW11}, which represent node and edge types between two neighboring nodes in a HG, have been proposed to model such rich information.  
	Formally, a path $v_1 \xrightarrow{R_1} v_2 \xrightarrow{R_2} \cdots \xrightarrow{R_{n-1}} v_n$
	is defined as a meta-path between nodes $v_1$ and $v_n$, wherein $R = R_1 \circ R_2 \circ \cdots \circ R_{n-1}$ defines the composite relations between node $v_1$ and $v_n$~\cite{DCS17}.
	In this paper, we intend to utilize symmetric and undirected meta paths to denote the closeness among target-type nodes $\mathcal{V}_t$, which can help simplify the problem setting. We represent the set of meta paths used in this paper as $\{\Phi_1, \Phi_2, \cdots, \Phi_P\}$, where $\Phi_i$ denotes the $i$-th meta path type. For example, in Figure~\ref{fig:acmhg}(d), Paper-Author-Paper (PAP) and Paper-Subject-Paper (PSP) are two types of meta-paths, which contain the semantic ``papers written by the same author" and ``papers belonging to the same subject" respectively.
	\noindent
	\begin{defn}[Meta-path based Adjacency Matrix]
		For the meta-path $\Phi_i$, if there exists a path instance between node $v_i \in \mathcal{V}_t$ and $v_j \in \mathcal{V}_t$, we call that $v_i$ and $v_j$ are ``connected neighbors'' based on $\Phi_i$. Such neighborhood information can be represented by a meta-path based adjacent matrix $A^{\Phi_i} \in \mathbb{R}^{|\mathcal{V}_t|\times |\mathcal{V}_t|}$, where $A_{ij}^{\Phi_i}=A_{ji}^{\Phi_i}=1$ if $v_i$, $v_j$ are connected by meta-path $\Phi_i$ and $A_{ij}^{\Phi_i}=A_{ji}^{\Phi_i}=0$ otherwise.
	\end{defn}
	%
	%
	
	
	In the next section, we will propose the novel method {\our} for heterogeneous graph representation learning, which is an unsupervised GNN model and can integrate information from different meta-paths through an attention mechanism.
	
	\begin{figure*}[t]
		\centering
		\includegraphics[width=1\textwidth]{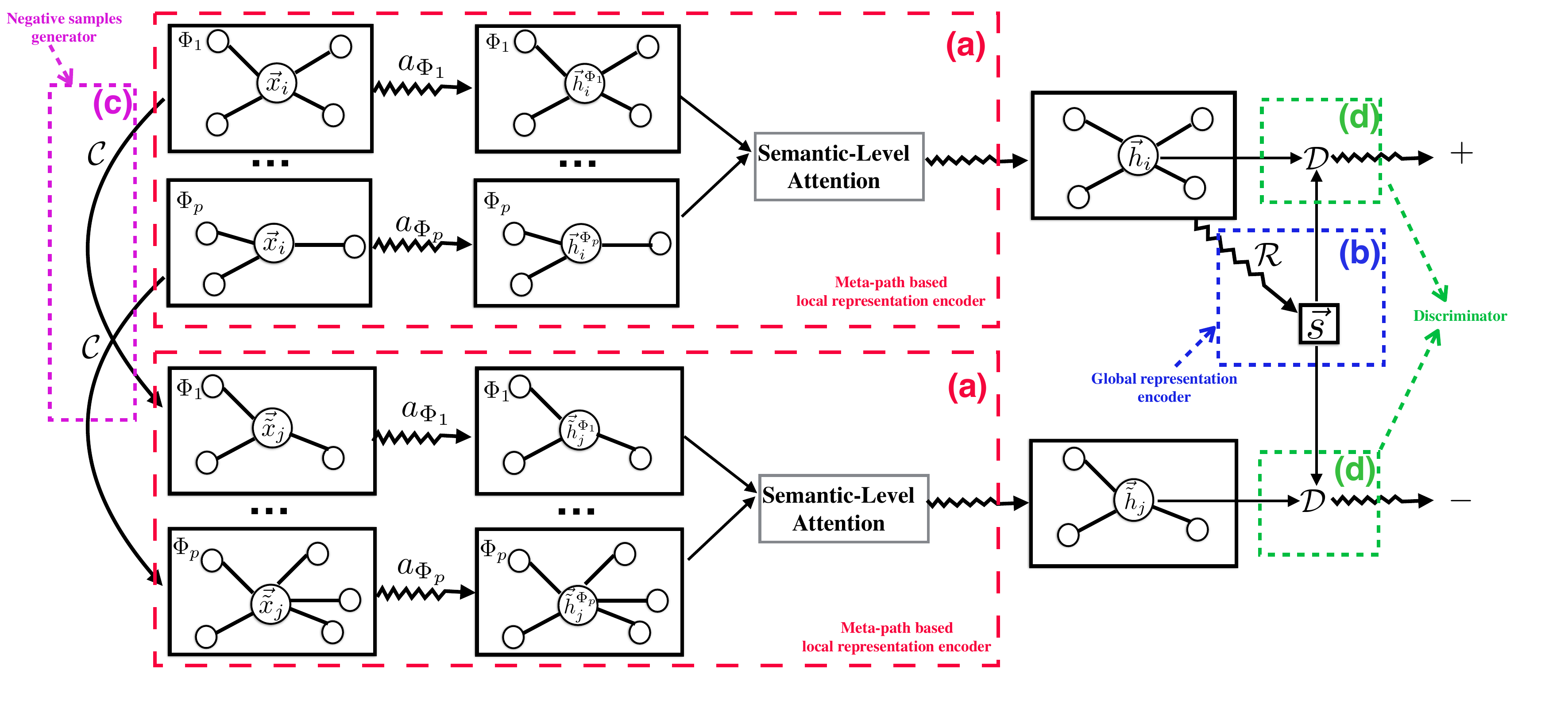}
		\captionsetup{singlelinecheck = false, format= hang, justification=raggedright, font=normalsize, labelsep=space}
		\caption{The high-level structure of {\our}. \textbf{(a)} Local representation encoder is a hierarchical structure: learning node representations in terms of every meta-path based adjacency matrix respectively and then aggregating them through semantic-level attention. \textbf{(b)} Global representation encoder $\mathcal{R}$ outputs a graph-level summary vector $\vec{s}$. \textbf{(c)} Negative samples generator $\mathcal{C}$ is responsible for generating negative nodes. \textbf{(d)} The discriminator $\mathcal{D}$ maximizes mutual information between positive nodes and the graph-level summary.}
		\label{fig:framework}
		
	\end{figure*}
	\section{{\our} Methodology}
	\label{sec:solution}


	\begin{table}[h]
		\centering
		\caption{Symbols and Definitions}
		\begin{tabular}{ll}
			\toprule
			Symbol & Interpretation  \\
			\midrule
			$\Phi$ & Meta-path\\
			$A^{\Phi}$ & Meta-path based adjacency matrix\\
			$X$ & The initial node feature matrix\\
			$\mathcal{V}_t$ & The set of nodes with the target type\\
			$N$ & The number of nodes in $\mathcal{V}_t$\\
			$\mathcal{G}$ & The given heterogeneous graph \\
			$\mathcal{D}$ & Mutual information based discriminator\\
			$\mathcal{C}$ & Negative samples generator\\
			$\mathcal{R}$ &Global representation encoder\\
			$\vec{s}$ & The graph-level summary vector \\
			$H^{\Phi}$ & Node-level representations \\
			$\vec{q}$ & Semantic-level attention vector \\
			$S^{\Phi}$ & Attention weight of meta-path $\phi$\\
			$\tilde{H}$ & Final negative nodes representations \\
			$H$ & Final positive nodes representations \\
			\bottomrule
			\label{tab:notation}
			\vspace{-20pt}
		\end{tabular}
	\end{table}
	\subsection{{\our} Architecture Overview}
	
	
	A high-level illustration of the proposed {\our} is shown in Figure~\ref{fig:framework}. 
	We summarize the notations used for model description in Table~\ref{tab:notation}.
	The input of {\our} is a heterogeneous graph $\mathcal{G}$ containing $N$ target-type nodes whose initial $d$-dimension features are denoted by $X\in \mathbb{R}^{N\times d}$, and meta-path set $\{{\Phi_i}\}_{i=1}^P$. Based on $\{{\Phi_i}\}_{i=1}^P$ we can calculate the meta-path set based adjacency matrices $\{A^{\Phi_i}\}_{i=1}^P$.
	The meta-path based local representation encoding described in Section~\ref{sec:local} has two steps: (1) learning individual node representation $H^{\Phi_i}$ from $X$ and each $A^{\Phi_i}, i=1,2..,P$ and (2) generating node representation $H$ by aggregating $\{H^{\Phi_i}\}_{i=1}^P$ through a semantic-level attention mechanism. A global representation encoder  $\mathcal{R}$ is proposed to derive a graph summary vector $\vec{s}$ from $H$ (see Section~\ref{sec:global}). The discriminator $\mathcal{D}$ will be trained with the objective to maximize mutual information between positive nodes and the graph-level summary $\vec{s}$. At the same time, {\our} is optimized in an end-to-end manner by backpropagation with the object of mutual information maximization. In Section~\ref{sec:learning} we elaborate the mutual information based discriminator $\mathcal{D}$ and the negative sample generator $\mathcal{C}$. 
	\subsection{Meta-path based local representation encoder}\label{sec:local}
	Meta-path based local representation encoder has a hierarchical structure. The first step is meta-path specific node representation learning, which encodes nodes in terms of each meta-path based adjacency matrix. The second step, namely heterogeneous graph node representation learning, is aggregating these representations relating to different meta-paths by a semantic-level attention mechanism.
	\subsubsection{Meta-path specific node representation learning}
	%
	We use meta-path based adjacency matrices to convey multiple different semantics due to the heterogeneity of input graphs. Given the meta-path $\Phi$, the adjacency matrix $\mathcal{A}^{\Phi} \in \mathbb{R}^{|\mathcal{V}_t|\times |\mathcal{V}_t|}$ can represent the relational information about the meta-path $\Phi$ based connections.
	Each of $A^{\Phi_i}, i=1,2,..P$ can be viewed as a homogeneous graph. 
	The initial node feature matrix $X \in \mathbb{R}^{N \times F}$ is constructed by stacking the feature vectors in $\mathcal{X} $. 
	At this step, our target is to derive a node representation containing the information of initial node feature $X$ and $A^{\Phi_i}$, with a node-level encoder:
	\begin{equation}
		H^{\Phi_i}=	a_{\Phi_i}(X, A^{\Phi_i})
	\end{equation}
	
	Two kinds of the encoder are considered in this work. The first is Graph Convolutional Network (GCN)~\cite{KW17}. 
	GCN introduces a spectral graph convolution operator for the graph representation learning. The node representations learned by GCN should be:
	\begin{equation}
		H^{\Phi_i} = ({D^{\Phi_i}}^{-\frac{1}{2}}\tilde{A^{\Phi_i}}{D^{\Phi_i}}^{-\frac{1}{2}})XW^{\Phi_i}
	\end{equation}
	where $\tilde{A^{\Phi_i}} = A^{\Phi_i} + I$, $D^{\Phi_i}$ is the diagonal node degree matrix of $\tilde{A^{\Phi_i}}$. Matrix $W^{\Phi_i} \in \mathbb{R}^{d \times F}$ is the filter parameter matrix, which is not shared between different $A^{\Phi}$. 
	
	The second encoder we consider is the Graph Attention Networks (GAT)~\cite{VCCRLB18}. GAT effectively updates the node representations by aggregating the information from their neighbors, including self-neighbor. For the $m$-th node, its $K$-head attention output can be computed as:
	
	\begin{equation}
		\vec{h}_m^{\Phi_i} = \concatenate_{k=1}^K \sigma(\sum_{j \in \mathcal{N}_{m}^{\Phi_i}}\alpha_{mj}^{\Phi_i,k}W^{\Phi_i} \vec{x}_j)
	\end{equation}
	where $\concatenate$ is the concatenation operator, $W^{\Phi_i}$ is the 
	linear transformation parameter matrix and $\mathcal{N}_{m}^{\Phi_i}$ is neighbor set defined by $\Phi_i$. $\alpha_{mj}^{\Phi_i,k}$ is the normalized attention coefficient computed by the $k$-th attention mechanism.
	
	After the meta-path specific node representation learning, we obtain the set of node representations $\{H^{\Phi_i}\}_{i=1}^P$. They are aggregated to get the heterogeneous graph node representation. 
	
	\subsubsection{Heterogeneous graph node representation learning}
	The representations learned based on the specific meta-path contain only the semantic-specific information.  In order to aggregate the more general representations of the nodes, we need to combine these representations $\{H^{\Phi_1}, H^{\Phi_2}, \dots, H^{\Phi_P}\}$.
	The vital issue to accomplish the aggregation is exploring how much each meta-path should contribute to the final representations. Here we add a semantic-level attention layer $L_{att}$ to learn the weights:
	\begin{equation}
		\{\beta^{\Phi_1},\beta^{\Phi_2},\dots,\beta^{\Phi_P}\} = L_{att}(H^{\Phi_1}, H^{\Phi_2}, \dots, H^{\Phi_P})
		\label{eq:model}
	\end{equation}
	Specifically, the importance of the meta-path $\Phi_i$ is calculated by
	\begin{equation}
		e^{\Phi_i} =\frac{1}{N} \sum_{n =1}^N \mathrm{tanh}(\vec{q}^{\mathrm{T}} \cdot [W_{sem}\cdot\vec{h}_n^{\Phi_i}+\vec{b}])
		\label{eq:importance}
	\end{equation}
	where $W_{sem}$ is a linear transformation parameter matrix. $\vec{q}$ is the learnable shared attention vector. $\beta^{\Phi_i}$ is obtained by normalizing $\{e^{\Phi_i}\}_{i=1}^P$ with a softmax function:
	\begin{equation}
		\beta^{\Phi_i} = \textup{softmax}(e^{\Phi_i}) = \frac{\mathrm{exp}(e^{\Phi_i})}{\sum_{j =1}^P \mathrm{exp}(e^{\Phi_j})}
	\end{equation}
	The heterogeneous graph node representation $H$ is obtained by a linear combination of $\{H^{\Phi_i}\}_{i=1}^P$, that is 
	\begin{equation}
		H = \sum_{i =1}^P \beta^{\Phi_i} \cdot H^{\Phi_i}
		\label{eq:embeddings}
	\end{equation}
	
	Our semantic attention layer is inspired by HAN~\cite{WJSWCYY19}, but there are still some differences in the learning direction. HAN utilizes classification cross-entropy as the loss function. Available labels in training set guide the learning direction. The attention weights learned in {\our} are guided by the binary cross-entropy loss, which indicates whether the node belongs to the original graph. Therefore, the weights learned in {\our} serve for the existence of a node. Because no classification label involves, the weights get no bias from the known labels.
	
	The representations $H$ serve as the final output local features. 
	It should be mentioned that all parameters in the meta-path based local representation encoder are shared for positive and negative nodes. Negative nodes are generated by the negative samples generator, which we will introduce in section~\ref{sec:nsg}. 
	The global representation encoder leverages the representations $H$ to output the graph-level summary, described in the following section.
	
	\subsection{Global Representation Encoder}\label{sec:global}
	The learning objective of {\our} is to maximize the mutual information between local representations and global representation. The local representations of nodes are included in $H$. We need the summary vector $\vec{s}$ to represent the entire heterogeneous graph's global information. Based on $H$, we examined three candidate encoder functions:
	
	\noindent
	\textbf{Averaging encoder function}. Our first candidate encoder function is the averaging operator, where we simply
	take the mean of the node representations to output the graph-level summary vector $\vec{s}$:
	
	\begin{equation}
		\vec{s} =\sigma\Bigg(\frac{1}{N} \sum_{i =1}^N \vec{h}_i\Bigg)
	\end{equation}
	$\sigma$ is a PReLU activation funtion.
	\noindent

	\noindent
	\textbf{Pooling encoder function}. In this pooling encoder function, each node's vector is independently fed through a fully-connected layer. An elementwise max-pooling operator has applied to summary the information from the nodes set:
	\begin{equation}
		\vec{s}_{pool} =max(\sigma(W_{pool}\vec{h}_i + b), i \in \{1,2,\dots,N\})
	\end{equation}
	\noindent
	where $max$ denotes the element-wise max operator and $\sigma$ is a nonlinear activation function.
	
	\noindent
	\textbf{Set2vec encoder function}. The final encoder function we examine is Set2vec \cite{VBK16}, which is based on an LSTM architecture. 
	Because the original set2vec in \cite{VBK16} works on ordered node sequences. However, we need a summary of the graph concluding comprehensive information from each node instead of merely graph structure. Therefore, we apply the LSTMs to a random permutation of the node's neighbor on an unordered set.
	
	\subsection{{\our} Learning}\label{sec:learning}
	\subsubsection{Mutual information based discriminator}
	It is inconvenient to calculate the mutual information between random variables directly. 
	Belghazi et al.~\cite{BBROBCD18} prove that the KL-divergence admits the Donsker-Varadhan representation and the $f$-divergence representation as dual representations. The dual representations provide a lower-bound to the mutual information of random variables $X$ and $Y$:
	\begin{equation}
		\text{MI}(X;Y) \geq \mathbb{E}_{\mathbb{P}_{XY}}[T_{\omega}(x,y)]-log(\mathbb{E}_{\mathbb{P}_{X} \otimes \mathbb{P}_{Y}}[e^{T_{\omega}(x,y)}])
	\end{equation}\label{eq:mi}	
	Here, $\mathbb{P}_{XY}$ is the joint distribution, and $\mathbb{P}_{X} \otimes \mathbb{P}_{Y}$ is the product of margins. $T_{\omega}$ is a deep neural network based discriminator parametrized by $\omega$. The expectations in equ.10 can be estimated using samples from $\mathbb{P}_{XY}$ and $\mathbb{P}_{X} \otimes \mathbb{P}_{Y}$. The expressive power of the discriminator ensures to approximate the \text{MI} with high accuracy. 
	Following the philosophy in \cite{HFMGBTB19}, we estimate and maximize the mutual information by training a discriminator $\mathcal{D}$ to distinguish positive sample set $Pos=\left\{[\vec{h}_n, \vec{s}]\right\}_{n=1}^N$ with negative sample set $Neg=\left\{[\vec{\tilde{h}}_m, \vec{s}]\right\}_{m=1}^M$. The sample $(\vec{h}_i, \vec{s})$ is denoted as positive as node $\vec{h}_i$ belongs to the original graph (the joint distribution), and $(\vec{\tilde{h}}_j, \vec{s})$ is negative as the node $\vec{\tilde{h}}_j$ is the generated fake one (the product of marginals). The discriminator $\mathcal{D}$ is a bilinear layer:

	\begin{equation}
		\mathcal{D}(\vec{h}_i, \vec{s}) = \sigma(\vec{h}_i^\mathrm{T}W_D\vec{s})
		\label{eq:discriminator}
	\end{equation}	
	Here $W_D$ is a learnable matrix, and $\sigma$ is the sigmoid activation function.
	Veli{\v{c}}kovi{\'c} et al.~\cite{velivckovic2018deep} prove that the binary cross-entropy loss amounts to maximizing the mutual information with theoretical guarantee.
	Therefore, we can maximize the mutual information with the binary cross-entropy loss of the discriminator:
	\begin{align}
		\mathcal{L}(Pos, Neg, \vec{s}) =  \frac{1}{N+M} \Bigg(&\sum_{n=1}^N \mathbb{E}_{Pos}[\mathrm{log} \mathcal{D}(\vec{h}_n, \vec{s})] \nonumber\\
		&+ \sum_{m=1}^M \mathbb{E}_{Neg}[\mathrm{log}(1-\mathcal{D}(\vec{\tilde{h}}_m, \vec{s}))]\Bigg) 
		\label{eq:loss}
	\end{align}		
	
	In essence, the discriminator works to maximize the mutual information between a high-level global representation and local representations (node-level), which encourages the encoder to learn the information presenting in all globally relevant locations. The information about a class label can be one of the cases. The above loss can be optimized through gradient descent. The representations of nodes can be learned when the optimization is completed. 
	
	\subsubsection{Negative samples generator}\label{sec:nsg}	
\begin{figure}[t]
		\centering
		\begin{minipage}[l]{1\columnwidth}
			\centering
			\includegraphics[width=\textwidth]{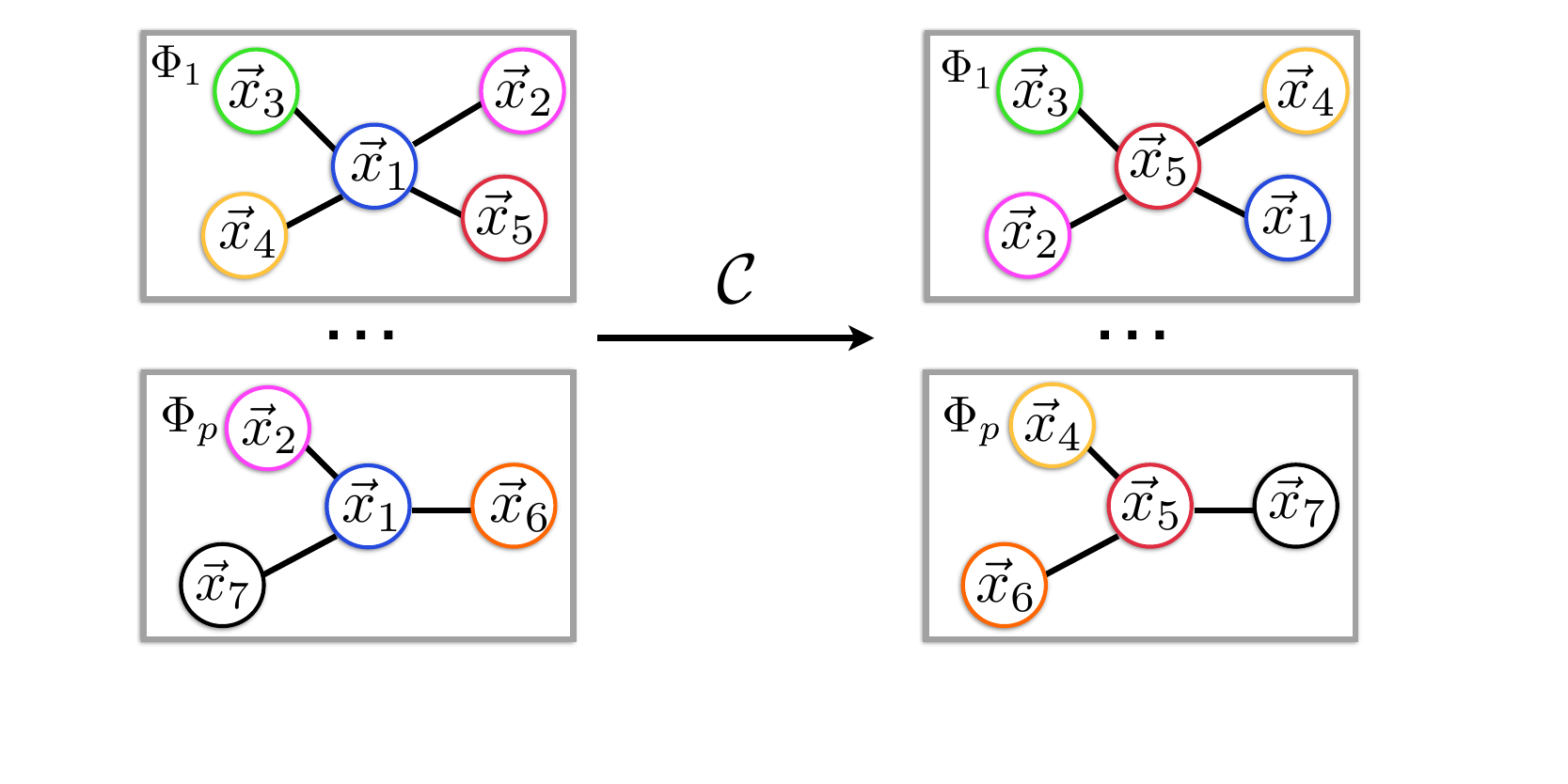}
		\end{minipage}
		\caption{The example of generating negative samples}\label{fig:shuffle}
	\end{figure}

	The negative sample set $\left\{[\vec{\tilde{h}}_m, \vec{s}]\right\}_{m=1}^M$ is composed of the samples that do not exist in the heterogeneous graph. As our target is to maximize the mutual information between positive nodes and the graph-level summary vector, the generated negative samples will affect the structural information captured by the model. In this way, we need high-quality negative samples that keep the structural information preciseness. We extend the negative sample generation approach proposed in~\cite{velivckovic2018deep} to heterogeneous graph setting. 
	In heterogeneous graph $\mathcal{G}$, we have rich and complex structural information characterized by meta-path based adjacency matrices. Our negative samples generator:
	
	\begin{equation}
		\tilde{X},\{A^{\Phi_1},A^{\Phi_2},\dots,A^{\Phi_P}\} = \mathcal{C}(X,\{A^{\Phi_1},A^{\Phi_2},\dots,A^{\Phi_P}\})
		\label{eq:corrupt}
	\end{equation}
	keeps all meta-path based adjacency matrices unchanged but shuffles the rows of the initial node feature matrix $X$, which changes the index of nodes to corrupt the node-level connections among them. 
	We provide a simple example to illustrate the procedure of generating negative samples in Figure~\ref{fig:shuffle}.

	
	\section{Evaluation}\label{sec:real-eval}
	
	%
	In this section, we evaluate the proposed \our\ framework in three real-world heterogeneous graphs. We first introduce the datasets and experimental settings. Then we report the model performance as compared to other state-of-the-art competitive methods. The evaluation results show the superiority of our developed model.

	\subsection{Datasets}
	We evaluate the performance of {\our} on three heterogeneous graphs and summarize their details in Table~\ref{tab:dataset}. 
	
	\begin{table}[h]
		\centering
		\caption{Statistics of experimented datasets.}
			    \tiny
		\begin{tabular}{l|ccccc|c}
			\toprule
			Dataset  & Node-type & \# Nodes & Edge-type & \# Edges & feature & Meta-path\\
			\midrule
			\multirow{3}{*}{ACM} & Paper (\textbf{P}) & 3025 & \multirow{1.8}{*}{Paper-Author} & \multirow{1.8}{*}{9744} & \multirow{1.8}{*}{1870} &  \multirow{1.8}{*}{PAP}  \\
			& Author (\textbf{A})  & 5835 & \multirow{1.8}{*}{Paper-Subject}  & \multirow{1.8}{*}{3025} & & \multirow{1.8}{*}{PSP} \\
			& Subject (\textbf{S})  & 56 &  & & & \\
			\midrule
			\multirow{4}{*}{IMDB} & Movie (\textbf{M}) & 4275 & \multirow{1.8}{*}{Movie-Actor} & \multirow{1.8}{*}{12838} &  &  \multirow{1.8}{*}{MAM} \\
			& Actor (\textbf{A}) & 5431 & \multirow{1.8}{*}{Movie-Director}  & \multirow{1.8}{*}{4280} & \multirow{2}{*}{6344}& \multirow{1.8}{*}{MDM}\\
			& Director (\textbf{D}) & 2082 &\multirow{1.8}{*}{Movie-keyword}  & \multirow{1.8}{*}{20529} & & \multirow{1.8}{*}{MKM}\\
			& Keyword (\textbf{K}) & 7313 &  &  & &\\
			\midrule
			\multirow{4}{*}{DBLP} & Author (\textbf{A}) & 4057 & \multirow{1.8}{*}{Author-Paper} & \multirow{1.8}{*}{19645} &  &  \multirow{1.8}{*}{APA} \\
			& Paper (\textbf{P}) & 14328 & \multirow{1.8}{*}{Paper-Conference}  & \multirow{1.8}{*}{14328}& \multirow{2}{*}{334} & \multirow{1.8}{*}{APCPA} \\
			& Conference (\textbf{C}) & 20 &\multirow{1.8}{*}{Paper-Term}  & \multirow{1.8}{*}{88420} & & \multirow{1.8}{*}{APTPA}\\
			& Term (\textbf{T}) & 8789 &  & & & \\
			\bottomrule
		\end{tabular}
		\label{tab:dataset}
	\end{table}
	
	\begin{itemize}[leftmargin=*]
		\item~DBLP~\cite{GLFSH09}: This is a research paper set, which contains scientific publications and the corresponding authors. The target author node can be divided into four areas: database, data mining, information retrieval, and machine learning. We use the area of authors as labels. The initial features are generated based on authors' profiles with the bag-of-words embeddings.
		The meta-paths we defined in DBLP are Author-Paper-Author (APA), Author-Paper-Conference-Paper-Author (APCPA), and Author-Paper-Term-Paper-Author (APTPA).
		\item~ACM~\cite{WJSWCYY19}: This is another academic paper data in which target paper nodes are categorized into 3 classes: database, wireless communication, and data Mining. We extract 2 meta-paths from this graph: Paper-Author-Paper (PAP) and Paper-Subject-Paper (PSP). 
		The initial features are constructed from paper keywords with the TF-IDF based embedding techniques.
		\item~IMDB~\cite{wu2016explaining}: It is a knowledge graph data about movies (target nodes) categorized into three types: Action, Comedy, and Drama. The meta-paths we choose are Movie-Actor-Movie (MAM), Movie-Director-Movie (MDM), and Movie-Keyword-Movie (MKM). 
		The feature of a movie is composed of \{color, title, language, keywords, country, rating, year\} with a TF-IDF encoding. 
		
	\end{itemize}

	\subsection{Experiment Setup}
	The most commonly used tasks to measure the quality of learned representations are node classification~\cite{PAS14,GL16,HYL17b} and node clustering~\cite{DCS17,WJSWCYY19}. We evaluate {\our} from both two types of tasks.
	\subsubsection{Compared Baselines}
	{\our} is compared with the following supervised and unsupervised methods:\\
	\noindent
	\textbf{\textit{Unsupervised methods}}
	\begin{itemize}
		\item~{Raw Feature}: The initial features are used as embeddings.
		\item~{Metapath2vec (M2V)~\cite{DCS17}}: A meta-path based graph embedding method for heterogeneous graph. We test all meta-paths and report the best result.
		\item~{DeepWalk (DW)~\cite{PAS14}}: A random walk based graph embedding method, but it is designed to deal with homogeneous graph. 
		\item~{DeepWalk+Raw Feature(DW+F)}: It concatenates the learned DeepWalk embeddings with the raw features as the final representations.
		\item~{DGI~\cite{velivckovic2018deep}}: A mutual information based graph representation method for homogeneous graph.
		\item~{HDGI-C}: The graph convolutional network is utilized to capture local representations in {\our}.
		\item~{HDGI-A}: This is another variant of {\our}, which uses graph attention mechanism to learn local representations.
	\end{itemize}
	\noindent
	\textbf{\textit{Supervised methods}}
	\begin{itemize}
		\item~{GCN~\cite{KW17}}: A semi-supervised methods for node classification on homogeneous graphs.
		\item~{RGCN~\cite{schlichtkrull2018modeling}}: It performs representation learning on all nodes labeled with entity types in heterogeneous graphs.
		\item~{GAT~\cite{VCCRLB18}}: GAT applies the attention mechanism in homogeneous graphs for node classification.
		\item~{HAN~\cite{WJSWCYY19}}: HAN employs node-level attention and semantic-level attention to capture the information from all meta-paths.
	\end{itemize}
	
	For methods designed for homogeneous graphs, i.e., DeepWalk, DGI, GCN, GAT, we do not consider graph heterogeneity and construct meta-path based adjacency matrix, then we report the best performance. We test all meta-paths for Metapath2vec and report the best result. For RGCN, because our task is to learn the representations of target-type nodes, the cross-entropy loss is calculated by the classification in target-type nodes only. In the node classification task, a training set is used to learn a simple classifier for unsupervised methods, while the supervised methods can output the result as end-to-end. For the node clustering, we will not use any label in this unsupervised learning task and make comparison among all unsupervised learning methods.

	\subsubsection{Reproducibility}
	For the proposed {\our}, including HDGI-C and HDGI-A, we optimize the model with Adam~\cite{KB15}. The dimension of node-level representations in HDGI-C is set as 512, and the dimension of $\vec{q}$ is set as 8. For HDGI-A, we set the dimension of node-level representations as 64, and the attention head is set as 4. The dimension of $\vec{q}$ is set as 8 as well. We employ Pytorch to implement our model and conduct experiments in the server with 4 GTX-1080ti GPUs. The detailed parameters of all other comparison methods are provided in the opensource codes:
	\href{https://github.com/YuxiangRen/Heterogeneous-Deep-Graph-Infomax}{https://github.com/YuxiangRen/Heterogeneous-Deep-Graph-Infomax.}  
	
\begin{table*}[thb!]
		\centering
		
		\begin{threeparttable}
			\caption{The results of node classification tasks}
			\begin{tabular}{l|c|c| c| c c|cccc| cccc }
				\toprule
				\multicolumn{3}{c}{\textbf{Available data}}&  \multicolumn{1}{c}{\textbf{X}}
				& \multicolumn{2}{c}{\textbf{A}}&\multicolumn{3}{c}{\textbf{X, A, Y}} &\multicolumn{4}{c}{\textbf{\ \ \ \ \ \ \ \ \ \ \ \ \ \ \ \ \ \ \ \ \ \ \ \ \ \ \ \ \ \ \ \ \ \ \ X, A}}  \\
				\cmidrule(l){1-14}
				Dataset  & Train & Metric & Raw & M2V & DW  & GCN & RGCN & GAT & HAN & DW+F & DGI & \textbf{HDGI-A} & \textbf{HDGI-C}\\
				\midrule
				\multirow{6}{*}{ACM} & \multirow{2}{*}{20\%} & Micro-F1 & 0.8590 & 0.6125 & 0.5503 & 0.9250 &0.5766 & 0.9178 & 0.9267 & 0.8785 & 0.9104 & 0.9178 & \textbf{0.9227} \\
				\cmidrule(l){3-14}
				& & Macro-F1 & 0.8585 & 0.6158 & 0.5582  & 0.9248 &0.5801 & 0.9172 & 0.9268 & 0.8789 & 0.9104 & 0.9170 & \textbf{0.9232}\\
				\cmidrule(l){2-14}
				& \multirow{2}{*}{80\%} & Micro-F1 & 0.8820 & 0.6378 & 0.5788  & 0.9317 & 0.5939 & 0.9250 & 0.9400 & 0.8965 & 0.9175 & 0.9333 & \textbf{0.9379}\\
				\cmidrule(l){3-14}
				&& Macro-F1 & 0.8802 & 0.6390 & 0.5825  & 0.9317 &0.5918 & 0.9248 & 0.9403 & 0.8960 & 0.9155 & 0.9330 & \textbf{0.9379}\\
				\midrule
				\multirow{6}{*}{DBLP} & \multirow{2}{*}{20\%} & Micro-F1 & 0.7552 & 0.6985 & 0.2805  & 0.8192 &0.1932 & 0.8244 & 0.8992 & 0.7163 & 0.8975 & 0.9062 & \textbf{0.9175}\\
				\cmidrule(l){3-14}
				& & Macro-F1 & 0.7473 & 0.6874 & 0.2302  & 0.8128 &0.2132 & 0.8148 & 0.8923 & 0.7063 & 0.8921 & 0.8988 & \textbf{0.9094}\\
				\cmidrule(l){2-14}
				& \multirow{2}{*}{80\%} & Micro-F1 & 0.8325 & 0.8211 & 0.3079  & 0.8383 &0.2175 & 0.8540 & 0.9100 & 0.7860 & 0.9150 & 0.9192 & \textbf{0.9226}\\
				\cmidrule(l){3-14}
				&& Macro-F1 & 0.8152 & 0.8014 & 0.2401  & 0.8308 &0.2212 & 0.8476 & 0.9055 & 0.7799 & 0.9052 & 0.9106 & \textbf{0.9153}\\
				\midrule
				\multirow{6}{*}{IMDB} & \multirow{2}{*}{20\%} & Micro-F1 & 0.5112 & 0.3985 & 0.3913 & 0.5931 &0.4350 & 0.5985 & 0.6077 & 0.5262 & 0.5728 & 0.5482 & \textbf{0.5893} \\
				\cmidrule(l){3-14}
				& & Macro-F1 & 0.5107 & 0.4012 & 0.3888 & 0.5869 &0.4468 & 0.5944 & 0.6027 & 0.5293 & 0.5690 & 0.5522 & \textbf{0.5914} \\
				\cmidrule(l){2-14}
				& \multirow{2}{*}{80\%} & Micro-F1 & 0.5900 & 0.4203 & 0.3953  & 0.6467 &0.4476& 0.6540 & 0.6600 & 0.6017 & 0.6003 & 0.5861 & \textbf{0.6592}\\
				\cmidrule(l){3-14}
				&& Macro-F1 & 0.5884 & 0.4119 & 0.4001  & 0.6457 &0.4527 & 0.6550 & 0.6586 & 0.6049 & 0.5950 & 0.5834 & \textbf{0.6646}\\
				\bottomrule
			\end{tabular}
			
			\begin{tablenotes}
				\item``\textbf{X}'' learning with initial features. ``\textbf{A}'' learning with the adjacency matrix. ``\textbf{Y}'' learning with node labels.
			\end{tablenotes}
		\end{threeparttable}\label{tab:classification_result}
\end{table*}
	\subsection{Performance Comparison}
	
	\subsubsection{Node classification task}
	In the node classification task, we train a logistic regression classifier for unsupervised learning methods, while the supervised methods output the classification result as end-to-end models. We conduct the experiments with two different training set sizes (20\% or 80\% of full datasets). The sizes of the validation set and test set are fixed at 10\% of full datasets. For unsupervised methods, the dataset division is used for training the logistic regression classifier but has no relationship with representation learning. To keep the results stable, we repeat the classification process 10 times and report the average Macro-F1 and Micro-F1 in Table~\ref{tab:classification_result}. \textbf{X}, \textbf{A} and \textbf{Y} in Table~\ref{tab:classification_result} denote initial features, the adjacency matrix and node labels, respectively. They are used to reflect the required input of different methods. We observe that HDGI-C outperforms all other unsupervised learning methods. When competing with the supervised learning methods (designed for homogeneous graphs like GCN and GAT), {\our} can perform much better. This observation proves that the type and semantic information are critical and need to be handled carefully instead of directly ignoring them in heterogeneous graphs. The result of RGCN is suboptimal because the original RGCN is a featureless approach, and we follow the code to assign a one-hot vector to each node.
	
	In addition, the unified learning of all types of nodes in the same latent space is beneficial to entity type classification. However, it may not be applicable to label classification. {\our} is also competitive with the result reported from HAN~\cite{WJSWCYY19}, which is designed for heterogeneous graphs. The reason should be that {\our} can capture more global structural information when exploring the mutual information in reconstructing the representation. At the same time, supervised loss based GNNs overemphasize the direct neighborhoods~\cite{velivckovic2018deep}. On the other hand, this observation suggests that the features learned through supervised learning in graph structures may have limitations, either from the structure or a task-based preference. These limitations can damage the representations from a more general perspective. 
	Both HDGI-C and HDGI-A achieve stunning performance, which verifies the effectiveness of the general framework of {\our} as well.
	\begin{table}
		\centering
		\caption{Evaluation results on the node clustering task}
		\begin{tabular}{l| c| c| c| c| c| c}
			\toprule
			Data & \multicolumn{2}{c|}{ACM} & \multicolumn{2}{c|}{DBLP} & \multicolumn{2}{c}{IMDB}\\
			\cmidrule(l){1-7}
			Method & NMI & ARI & NMI & ARI & NMI & ARI \\
			\cmidrule(l){1-7}
			DeepWalk    &25.47  & 18.24 & 7.40  & 5.30 & 1.23 & 1.22 \\
			Raw Feature &32.62  & 30.99 & 11.21 & 6.98 & 1.06 & 1.17 \\
			DeepWalk+F  &32.54  & 31.20 & 11.98 & 6.99 & 1.23 & 1.22 \\
			Metapath2vec&27.59  & 24.57 & 34.30 & 37.54 & 1.15 & 1.51 \\
			DGI         &41.09  & 34.27 & 59.23 & 61.85 & 0.56 & 2.6 \\
			\cmidrule(l){1-7}
			HDGI-A      &\textbf{57.05} & \textbf{50.86} & 52.12 & 49.86 & 0.8 & 1.29 \\
			HDGI-C      & 54.35 & 49.48 & \textbf{60.76} & \textbf{62.67} & \textbf{1.87} & \textbf{3.7} \\
			\bottomrule
		\end{tabular}
		\label{tab:results_Rossmann}
	\end{table}
	\subsubsection{Node clustering task}
	\begin{figure}[t]
			\centering
		\subfigure[Macro-F1]{\label{fig:Macro_F1}
			\begin{minipage}[l]{0.8\columnwidth}
				\centering
				\includegraphics[width=1\textwidth]{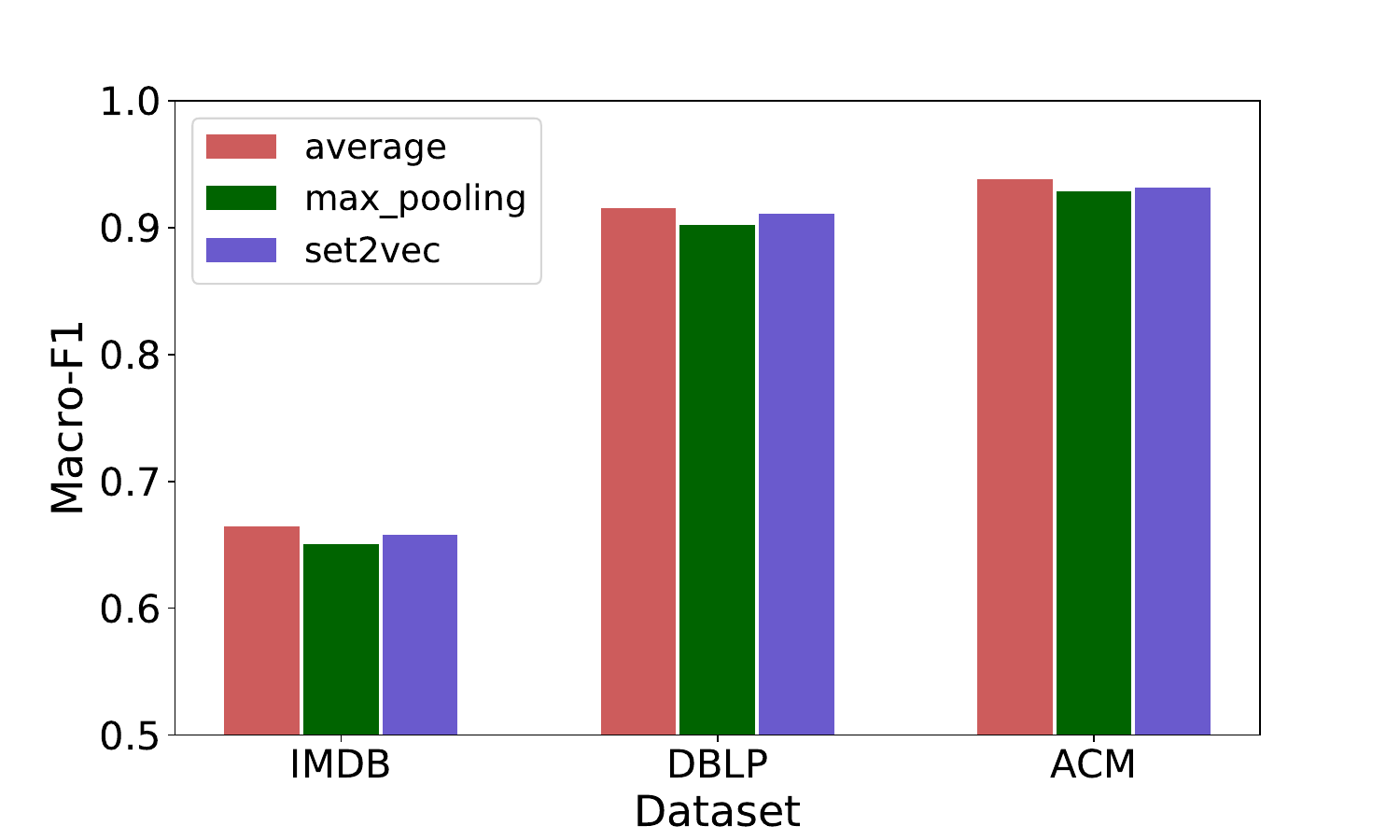}\vspace{3pt}
			\end{minipage}
		}
		\subfigure[Micro-F1]{\label{fig:Micro_F1}
			\begin{minipage}[l]{1\columnwidth}
				\centering
				\includegraphics[width=0.8\textwidth]{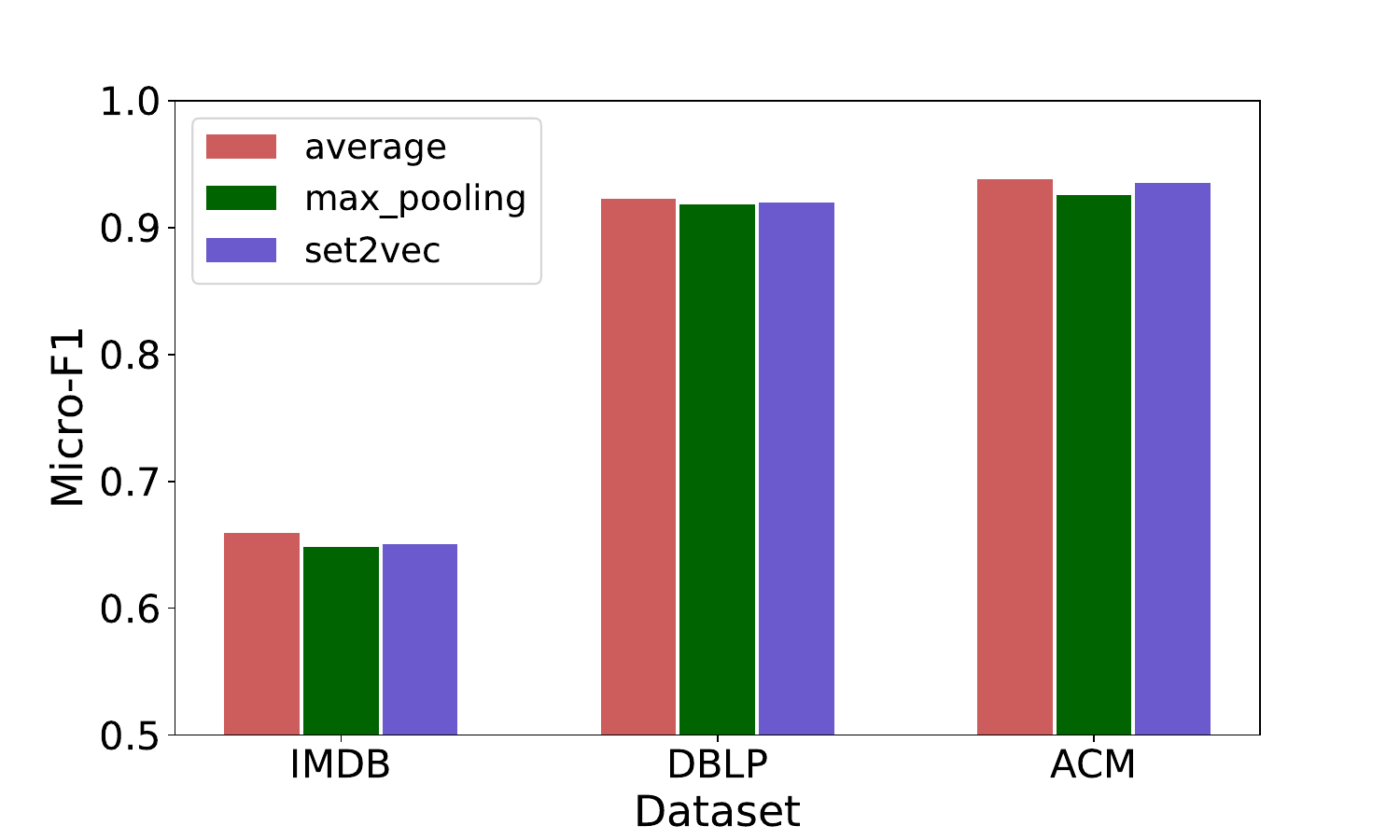}\vspace{3pt}
			\end{minipage}
		}
		\caption{The comparison between different global representation encoder functions}\label{fig:global_function}
	\end{figure}
	In the node clustering task, we use the K-Means to conduct the clustering based on the learned representations. The number of clusters $K$ is set as the number of target node classes. We do not use any label in this unsupervised learning task and make the comparison among all unsupervised learning methods. We also repeat the clustering process 10 times and report the average NMI and ARI in Table~\ref{tab:results_Rossmann}. DeepWalk cannot perform well because they are not able to handle the graph heterogeneity. Metapath2vec cannot handle diversified semantic information simultaneously, which makes the representations not effective enough. The verification based on node clustering tasks also demonstrates that {\our} can learn effective representations by considering the structural information, the semantic information, and the node independent content simultaneously.
	
	\subsubsection{HDGI-A vs HDGI-C}
	From the comparison between HDGI-C and HDGI-A in node classification tasks, the results reflect some interesting findings. HDGI-C achieves better performance than HDGI-A in all experiments, which means that the graph convolution works better than the attention mechanism in capturing local network structures. The reason might be that the graph attention mechanism is strictly limited to the direct neighbors of nodes. The graph convolution considering hierarchical dependencies can see farther. The results of the clustering task can also verify this analysis.
	\subsubsection{Different global representation encoder functions}
	We present the results of HDGI-C with different global representation encoder functions working on the node classification task in Figure~\ref{fig:global_function}. The simple average function performs the best. However, we can find that this advantage is very subtle. In fact, each function can perform well on experimental datasets. Nevertheless, for larger and more complex heterogeneous graphs, a specified and sophisticated function may perform better. The design of the global encoder function for heterogeneous graphs with different scales and structures is an open question, which is worthy of further discussion.
	
	\section{Conclusion}
	\label{sec:conclusion}
	In this paper, we propose an unsupervised graph neural network, {\our}, which learns node representations in heterogeneous graphs. {\our} combines several state-of-the-art techniques. It employs convolution style GNNs along with a semantic-level attention mechanism to capture individual local representations of nodes. Through maximizing the local-global mutual information, {\our} learns high-level representations containing graph-level structural information. It exploits the structure of meta-path to model the connection semantics in heterogeneous graphs. Node attributes are fused into representations through the local-global mutual information maximization simultaneously. 
	We demonstrate the effectiveness of learned representations for both node classification and clustering tasks on three heterogeneous graphs. {\our} is incredibly competitive in node classification tasks with state-of-the-art supervised methods, where they have the additional supervised label information. We are optimistic that mutual information maximization is a promising future direction for unsupervised representation learning.
	
	\bibliographystyle{plain}
	\bibliography{refs}

\end{document}